\documentclass{article}
\usepackage{arxiv,natbib}
\usepackage{hyperref}
\usepackage{amsmath,amsfonts}
\usepackage[noend]{algorithmic}
\usepackage{graphicx,xcolor}
\usepackage{algorithm}
\usepackage{multirow,url}
\usepackage{lscape}

\setcitestyle{round}

\usepackage{tikz}
\usetikzlibrary{shapes.geometric, arrows}
\tikzstyle{startstop} = [rectangle, rounded corners, minimum width=1cm, minimum height=0.75cm, text centered, draw=black]
\tikzstyle{io} = [trapezium, trapezium stretches=true, trapezium left angle=70, trapezium right angle=110, minimum width=0.75cm, minimum height=0.5cm, text centered, draw=black]
\tikzstyle{process} = [rectangle, minimum width=1.5cm, minimum height=0.75cm, text centered, text width=1.5cm, draw=black]
\tikzstyle{algo} = [rectangle, minimum width=3.5cm, minimum height=1cm, text width=3.5cm, draw=black]
\tikzstyle{decision} = [diamond, minimum width=2cm, minimum height=0.5cm, text centered, draw=black]
\tikzstyle{arrow} = [thick,->,>=stealth]

\newcommand{\ignore}[1]{}

\begin{document}

\title{\bf Evolutionary Algorithm for Chance Constrained Quadratic Multiple Knapsack Problem}

\author{
        {Kokila Kasuni Perera} \\
        Optimisation and Logistics\\
	School of Computer and Mathematical Sciences\\
        The University of Adelaide\\
        Adelaide, Australia \\
        \And
	{Aneta Neumann} \\
 Optimisation and Logistics\\
	School of Computer and Mathematical Sciences\\
        The University of Adelaide\\
        Adelaide, Australia \\
}
\maketitle

\begin{abstract}

Quadratic multiple knapsack problem (QMKP) is a combinatorial optimisation problem characterised by multiple weight capacity constraints and a profit function that combines linear and quadratic profits. We study a stochastic variant of this problem where profits are considered as random variables. This problem reflects complex resource allocation problems in real-world scenarios where randomness is inherent. We model this problem using chance constraints to capture the stochastic profits. We propose a hybrid approach for this problem, which combines an evolutionary algorithm (EA) with a local optimisation strategy inspired by multi-factorial optimisation (MFO). EAs are used for global search due to their effectiveness in handling large, complex solution spaces. In the hybrid approach, EA periodically passes interim solutions to the local optimiser for refinement. The local optimiser applies MFO principles, which are typically used in multi-tasking problems. The local optimiser models the local problem as a multi-tasking problem by constructing disjoint search spaces for each knapsack based on an input solution. 
For each item, its assignment across all knapsacks is considered to determine the preferred knapsack. Items are then divided into disjoint groups corresponding to each knapsack, allowing each knapsack to be treated as a separate optimisation task. This structure enables effective application of MFO-based local refinements.
We consider two EAs for the problem, (1+1) EA and ($\mu+\lambda$) EA. We conduct experiments to explore the effectiveness of these EAs on their own and also with the proposed local optimiser. Experimental results suggest that hybrid approaches, particularly those incorporating MFO, perform well on instances where chance constraints and capacity constraints are tight.
\end{abstract}

\keywords{
quadratic multiple knapsack problem, chance constraints, evolutionary strategies, multi-factorial optimisation
}

\section{Introduction}

The quadratic multiple knapsack problem (QMKP) is a complex optimisation problem that belongs to the class of knapsack problems. Compared to the classical knapsack problem, QMKP is characterised by a non-linear objective function and multiple constraints, one for each knapsack \citep{CACCHIANI2022105693}. 
This problem generalises the classical knapsack, introducing $m$ knapsacks, each with its own capacity, and additionally includes a quadratic profit component. This additional profit accounts for the extra value generated by specific pairs of items assigned to the same knapsack. The presence of multiple knapsacks and the quadratic profit function makes this problem a challenging combinatorial optimisation problem. 
We can also view this problem as a combination of two well-known knapsack problem variants: the multiple knapsack problem (MKP) and the quadratic knapsack problem (QKP), both of which are NP-hard problems. Thus, it implies that QKMP is a strongly NP-hard problem \citep{GALLI202336}. The structure of QMKP reflects real-world problem scenarios that involve multiple constraints and interactions between selected items, contributing to overall value generation. This is applicable in various fields, including manufacturing \citep{SARAC201478}, project management, logistics, and capital budgeting.

In this study, we consider a stochastic variation of QMKP where item profits are uncertain. Uncertainties are unavoidable in real-world problem scenarios due to the inherent stochastic nature within the problem or its environment. The stochastic variation of QMKP with stochastic profits reflects real-world scenarios where it is challenging to determine objective values and discover reliable solutions. Studying this problem is crucial to identifying methods that generate reliable solutions for complex combinatorial optimisation problems under uncertainty. Motivated by recent research into knapsack problems with stochastic profits \citep{10.1007/978-3-031-70055-2_8,10.1007/978-3-031-14714-2_21,10.1145/3638529.3654081}, we introduce a chance-constrained model for stochastic QMKP. This model captures the uncertainty of profits by associating a required confidence level with the expected profit. Based on this, we propose a fitness function that aims to maximise profit while holding the required confidence level. This model incorporates the stochastic nature of profits into the non-linear objective function of QMKP through chance constraints. 

Evolutionary algorithms have been effective in addressing a range of combinatorial optimisation problems, particularly for those with chance constraints \citep{10.1007/978-3-031-70055-2_8,10.1145/3638529.3654081}. In this study, we explore both single-solution and population-based evolutionary strategies for stochastic QMKP. Furthermore, we are particularly interested in enhancing their performance by integrating local optimisation heuristics. We explore hybrid methods for the target problem, which combine global search via evolutionary optimisation with local optimisers focused on refining input solutions through small, problem-specific adjustments \citep{RYAN2003673}. 

In this study, we propose a local optimiser for stochastic QMKP, which treats each knapsack as an individual optimisation task. Decomposing the local problem into multiple optimisation tasks enables us to address the problem using multi-tasking optimisation approaches. We aim to adopt a multi-factorial evolutionary algorithm (MFEA), an evolutionary approach that addresses multiple optimisation tasks in a unified search process \citep{MFEA_Gupta_7161358}. Compared to other multi-task optimisation methods, multi-factorial optimisation (MFO) techniques are effective when the optimisation tasks share a related or similar structure. Although MFO naturally assumes the tasks to have disjoint search spaces, the search space of QMKP is the set of all items, and individual knapsacks share these items. Thus, this problem does not adhere to the above assumption. It requires adopting a problem model that allows one to apply MFO on QMKP as a local optimiser. To refine a given solution, we construct disjoint groups of items based on their preferences for the knapsack relative to the input solution. These preferences guide the mutation operator in generating offspring solutions that are in the neighbourhood of the input solution. Additionally, when MFO transfers knowledge between solutions, these disjoint subgroups help preserve qualities inherited from the input solution. This overall design supports the effective use of MFO for local improvements in stochastic QMKP, leveraging the individual knapsacks that compose the problem. 



\subsection{Background}

Since the earliest literature on QMKP, evolutionary strategies have been suggested to address the problem. QMKP was first introduced in \citet{3heuristics_QMKP_10.1145/1143997.1144096}, where three heuristic methods, including a greedy heuristic, a stochastic hill-climber method, and a genetic algorithm (GA), were proposed for the problem. The GA proposed in \citet{3heuristics_QMKP_10.1145/1143997.1144096} had limitations when addressing instances with larger problem sizes, whereas the hill-climber performed well as the number of elements and/or knapsacks grew. Later, another GA was proposed in \citep{GA_QMKP_10.1007/978-3-540-75555-5_47} that performs well for large problem instances. The latter algorithm achieves its performance by utilising variation operators designed to maintain the feasibility of a solution. An exact method for QMKP was proposed for the first time by \citet{ExactMethod_QMKP_10.1287/ijoc.2018.0840}, which is based on a branch-and-price approach. The recent study by \citet{GALLI202336} proposes mathematical methods for the problem, based on Lagrangian relaxation.

\citet{memeticQMKP} proposed a hybrid evolutionary approach for the deterministic QMKP for the first time to the best of our knowledge. They combined a link adjustment approach with evolutionary optimisation strategies where the optimisation relies on a powerful initialisation method and specific variation operators. 

Multi-factorial optimisation (MFO) is an evolutionary computing approach that is designed to simultaneously address problems with multiple optimisation tasks. MFO solves multiple tasks simultaneously in a shared search process, allowing it to transfer knowledge between them. \citep{MFEA_Gupta_7161358} introduced a key MFO algorithm, the multi-factorial evolutionary algorithm (MFEA), which unifies the optimisation tasks within a common search space. This setup enables MFEA to utilise a shared solution representation and manage solution variations using evolutionary optimisation techniques. MFO works well when the tasks are related or share some structure, and helps find reasonable solutions faster \citep{MFEA_Gupta_7161358}.

Among various approaches considered for deterministic QMKP, we have not identified any study that utilises a multi-task approach to consider individual knapsack goals. In particular, no previous research has explored local optimisation by decomposing the problem into individual optimisation tasks with respect to each knapsack and approaching the problem using a multi-task optimisation strategy. A popular method for multi-tasking optimisation is multi-factorial evolutionary optimisation (MFO). MFO is applicable when the optimisation tasks share the same or very similar structure, which allows it to use a common solution representation and use simple evolutionary operators. 


This paper investigates the effectiveness of evolutionary optimisation methods for QMKP where profits are stochastic. We model this problem as a chance-constrained problem to capture the uncertainties in item and pairwise profits. This model enables us to find reliable solutions for the problem under uncertain conditions. We propose using EAs to address the chance-constrained problem variant of QMKP. To further enhance EAs, we propose a preference-based local optimiser that uses MFO principles to improve performance. We combine this local optimiser with the simple $(1+1)$ EA as well as the population-based ($\mu+\lambda$) EA. The proposed local optimiser takes input from the EA and makes local improvements to it. The local optimisation identifies preferred knapsacks for each item relative to a reference solution and considers individual optimisation tasks for each knapsack. Then, a preference-based mutation operator is applied to evolve the reference solution, and MFO is used to optimise the profit of individual knapsacks. We experimentally evaluate the performance of selected EA on stochastic QMKP both with and without the proposed local optimiser. We conduct experiments on a set of benchmark QMKP instances under numerous chance-constrained settings to assess the effectiveness of the evolutionary methods. 

The rest of this paper is structured as follows. The next section, Section \ref{sec:problem}, provides a formal definition of the target problem, stochastic QMKP. Section \ref{sec:methods} describes the algorithms selected for the problem and proposed methods. Section \ref{sec:experiments} and \ref{sec:results} present the experimental setups and discuss their results, respectively. Finally, Section \ref{sec:conclusion} summarises the key aspects of the paper and provides concluding remarks.

\section{Profit Chance Constrained Quadratic Multiple Knapsack Problem}\label{sec:problem}


In this section, we introduce the target problem considered in this study, profit-chance constrained QMKP. First, we briefly introduce the classical deterministic QMKP. Then, we introduce its stochastic variant, which incorporates uncertain profits and a chance-constrained formulation of this target problem.

The standard QMKP problem is defined as follows. Given a set of elements $N=\{1,\ldots,n\}$, each element $i\in N$ is associated with a weight $w_i$ and a linear profit $p_i$. Additionally, each pair of elements $i,j\in N$ ($i\neq j$) yields a joint profit value $p_{i,j}$, which is commonly referred to as the quadratic or pairwise profit. Let $m$ be the number of knapsacks and define the set of knapsacks as $M=\{1,\ldots,m\}$, each knapsack $k\in M$ having a capacity of $C_k$. 

A solution to the problem $X=\{I_1, \ldots, I_m\}$ is a set of mutually exclusive subsets $I_1, \ldots, I_m$, where $I_k$ gives the set of elements assigned to knapsack $k\in M$ and $I_0$ denotes the set of items that are not chosen in the packing plan. The objective of QMKP is to find a solution, $X^*$, that maximises the sum of item profit and pairwise profit while satisfying the weight capacity constraint on each knapsack. 

We can formally present QMKP as follows, 
\begin{equation}
    \label{eq:prob_def}
    \begin{array}{c@{}c@{}c}
    \max  &{} \sum_{k\in M} \left( \sum_{i\in I_k} p_i + \sum_{i,j\in I_k,i\neq j} p_{ij} \right) &{}. \\
    \text {subject~to} 
    &{}  \sum_{i\in I_k} w_i < C_k 	&{}	\forall{k\in M}\\ 
    \text {and}   & I_k \cap I_l = \emptyset  			&{}	\forall{k,l\in M}, k\neq l\\ 
    \text {and}   & I_k \subset N  &{}	\forall{k\in M}
    \end{array}
\end{equation}


In the stochastic setting, we assume that both profits: linear profit and quadratic (pairwise) profit of QMKP elements, are stochastic. Since these profits are stochastic, it is not possible to determine the exact profit yields from a solution unless we assume a normal distribution for the profits. According to the literature on knapsack problems with stochastic profits, we can model the situation using chance constraints and derive chance-constrained profits using suitable tail-bound inequalities. 

Let $P^k$ denote the profit of the knapsack $k$. We associate a confidence level $\alpha(>0.5)$ with $P^k$, where $k \in M = \{1,\ldots,m\}$ to capture the effects of uncertain profits. Then, we reformulate the deterministic QMKP with chance constraints as follows,

\begin{equation}
\label{eq:cc_qmkp}
\begin{array}{c@{}c@{}l}
   \max &{} \sum_{k\in M} P^{k} \\
  \text {subject~to } &{} {Pr\left(\sum_{i\in I_k} p_i + \sum_{i,j\in I_k,i\neq j} p_{ij} \geq P^k\right)\geq \alpha} &\forall{k\in M}
  \\
  \text {and} &{} \sum_{i\in I_k} w_i < C_k &{}\forall{k\in M}\\ 
  \text {and} &{} I_k \cap I_l = \emptyset &{}\forall{k,l\in M}, k\neq l\\ 
  \text {and} &{} I_k \subset N &{}\forall{k\in M}
	
\end{array}
\end{equation}

This formulation ensures that the total profit of each knapsack $i$ holds the maximal value $P^i$ with at least $\alpha$ confidence. 

Next, we need to derive profit estimation for $P^i$, the profit of knapsack $i\in M$. \citet{10.1007/978-3-031-14714-2_21} defines the chance-constrained profit for a single knapsack problem with stochastic item profits. We consider their profit estimation defined based on the Chebyshev inequality. 
Given that $\mu(x)$ and $v(x)$ are the expectation and variance of profit of $x$, the profit of $x$ with the confidence level $\alpha = 1-\alpha^\prime$ is estimated as follows, (for proof, we refer the reader to \citet{10.1007/978-3-031-14714-2_21}), 
\begin{equation}
	\label{eq:kp_profit}
	\mu(x) - \sqrt{\frac{\alpha}{1-\alpha}} \cdot \sqrt{v(x)}
\end{equation}

In QMKP, the total profit of each knapsack comprises two independent random variables: the item profit $p_item$ and pairwise profit $p_pair$. Therefore, essentially, we define the expected value and variance for the sum of these two variables. Let $\mu_{item}$ and $\mu_{pair}$ denote the expected value of item and pairwise profits, and $v_{item}$ and $v_{pair}$ denote their variances. Since the two profit variables are independent, their covariance is 0. Thus, for a given solution $x$ of QMKP, we have, expected profit ($\mu(x)$) and variance ($v(x)$) as follows,

$$\mu(x) = \mu_{item}(x) + \mu_{pair}(x) $$
$$v(x) = v_{item}(x) + v_{pair}(x) $$

Then, we derive a profit estimate for the total item profits and pairwise profits from Equation \ref{eq:kp_profit}. This estimate gives us the profits of knapsack $i$, under the chance constraint as follows,

\begin{equation}
	\label{eq:qmkp_kp_profit}
	P^i_{Cheb} = \left(\mu_{item}(x) + \mu_{pair}(x)\right) - \sqrt{\frac{(1-\alpha)}{\alpha}} \cdot \sqrt{v_{item}(x) + v_{pair}(x)}
\end{equation}

We can use this estimation to evaluate the profit of each knapsack, which adds up to give the total profit of the solution. As this profit evaluation is derived from Chebyshev's inequality, it is applicable when the expected value and variance of item and pair profits are known or can be estimated.

\section{Methods}\label{sec:methods}
This section introduces the methods that are considered for the target problem in this paper. We mainly consider evolutionary algorithms for the stochastic QMKP. We describe the solution representation and evolutionary operators utilised in the EAs and the special adaptation of MFO methods for the problem.

\subsection{Problem Representation and Mutation Operations}\label{sec:representation_mutation}
We consider several evolutionary approaches to address the problem. When applying evolutionary algorithms, the solution representation is a crucial aspect, as it can restrict the use of certain evolutionary operators.

A solution to QMKP is several subsets of items ($I_k \subset N$) assigned to each knapsack. In this study, we represent such a solution using an integer sequence of length $n$, equal to the total number of items ($n$). Let $x=\{0,1, \ldots, m\}^n$ be a solution for QMKP. If the item $i$ is not assigned to any knapsack, then the $i$th integer in $x$ is set to $0$; otherwise, it is set to $1, \ldots, m$ depending on which knapsack it is assigned to. We consider this representation across all methods in this paper. 

The variation operators for the methods are selected and designed to comply with this representation. In this work, we consider two mutation operators for the problem. First, we consider the random resetting mutation operator, which is the equivalent of the bit-flip mutation operator adapted for integer strings. Given an integer sequence $x=\{0,1, \ldots, m\}^n$, each integer in $x$ is reset with $1/n$ probability to one of the states, $\{0,1, \ldots, m\}$ chosen uniformly at random. Secondly, we consider the swap mutation operator, which swaps the assigned state of two randomly selected integers in the sequence with each other. This mutation mimics the natural process of exchanging items between the knapsacks, and our preliminary investigations show that using both mutations in the optimisation improves the overall outcome of the algorithm. The next subsection describes how these two mutation operators are incorporated into evolutionary methods in this study.

\subsection{Evolutionary Algorithms for stochastic QMKP}

This study considers single-objective evolutionary approaches to address QMKP with a chance constraint on profit. We consider adaptations of the standard $(1+1)$ EA and $(\mu+\lambda)$ EA for the chance constrained QMKP variation targeted in this study. 

$(1+1)$ EA is the simplest form of an EA, which maintains a single solution and varies this solution to generate a single offspring in each iteration, which may replace the parent solution based on its fitness. The standard $(1+1)$ EA applies to problems where solutions are represented as bit strings and uses the bit-flip mutation to generate offspring solutions. Since we use an integer representation for QMKP solutions, we replace the bit-flip mutation with the two selected mutation operators, random resetting and swap mutation. The algorithms choose one of the two mutation operators with an equal likelihood and use it to generate offspring solutions. 

We also consider the population-based evolutionary approach $(\mu+\lambda)$ EA for the target problem. 
It is a generalisation of the $(1+1)$ algorithm, where it maintains a population of $\mu > 1$ solutions as the parent population and optimises considering $\lambda > 1$ offspring solutions in each generation. Both parents and offspring are considered together when selecting the best $\mu$ solutions for the next iteration. When generating offspring solutions, mutation operators are used in a similar manner to that in (1+1) EA. It randomly chooses between the mutation operators —random resetting and swap mutation —with equal likelihood and applies the selected operator to the parent solution to generate an offspring solution. Typically, $(\mu+\lambda)$ EA also applies crossover variations to solutions. For fair comparison between $(1+1)$ and $(\mu+\lambda)$ EA, we do not only consider mutation-based variations to generate offspring solutions.

\subsection{A Hybrid Evolutionary Approach for Stochastic QMKP}

In this section, we propose to use a hybrid evolutionary approach for the target problem, where local optimisation is utilised to enhance the results generated by EAs. Stochastic QMKP is a complex problem which involves uncertainties, multiple constraints and a complex objective function. Thus, simple randomised approaches can use local search heuristics to navigate the complex search space and efficiently identify high-quality solutions. 

We follow a simple hybrid approach where the optimisation alternates between the global and local optimisers after regular time intervals determined by the count of fitness evaluations. The outline of the proposed hybrid search approach is given in Figure \ref{fig:memetic_algo}. We utilise this method for QMKP as follows. The evolutionary global optimiser and local optimiser take turns to optimise the problem for a given portion of the total evaluation budget. The input of the local optimiser is the best solution observed by the global optimiser. Then, after local improvements are made to the input solution, a set of refined solutions is passed back to the global optimiser. The two algorithms take turns to optimise the problem, sharing the latest observations. 

We can use both (1+1) EA and ($\mu+\lambda$) EA as the global optimiser in this hybrid approach. Algorithm \ref{algo:1p1-mfo} and \ref{algo:muPlambda-mfo} show the steps of the hybrid approach when used with (1+1) EA and ($\mu+\lambda$) EA, respectively.

\begin{figure}[!t]
\centering
\begin{tikzpicture}[node distance=2cm]

\node (start) [startstop] {Start};

\node (nullSol) [process, right of=start]{$x_0=\emptyset$};
\node (in1) [io, right of=nullSol] {$x_0$};
\node (local) [algo, right of=in1, xshift=2.25cm] 
{\textbf{Local Optimiser:}\\	Refine $x_0$\\Output: optimised population};
\node (out1) [io, below of=local] {$P^\prime$};
\node (global) [algo, below of=out1] {\textbf{Global Optimiser (EA)}\\
Add $P^\prime$ and optimise.\\
Output: best solution};
\node (out2) [io, left of=global, xshift=-0.75cm] {$x^*$};
\node (dec1) [decision, left of=global, xshift=-2.75cm] {Terminate?};
\node (in2)[process, above of=dec1, yshift=0.25cm]{$x_0=x^*$};
\node (stop) [startstop, left of=dec1, xshift=-0.5cm] {Stop};

\draw [arrow] (start) -- (nullSol);
\draw [arrow] (nullSol) -- (in1);
\draw [arrow] (in1) -- (local);
\draw [arrow] (local) -- (out1);
\draw [arrow] (out1) -- (global);
\draw [arrow] (global) -- (out2);
\draw [arrow] (out2) -- (dec1);
\draw [arrow] (dec1) -- node[anchor=west] {no} (in2);
\draw [arrow] (in2) -- (in1);
\draw [arrow] (dec1) -- node[anchor=south] {yes} (stop);

\end{tikzpicture}
\caption{Outline of Hybrid Evolutionary Optimisation Approach}\label{fig:memetic_algo}
\end{figure}
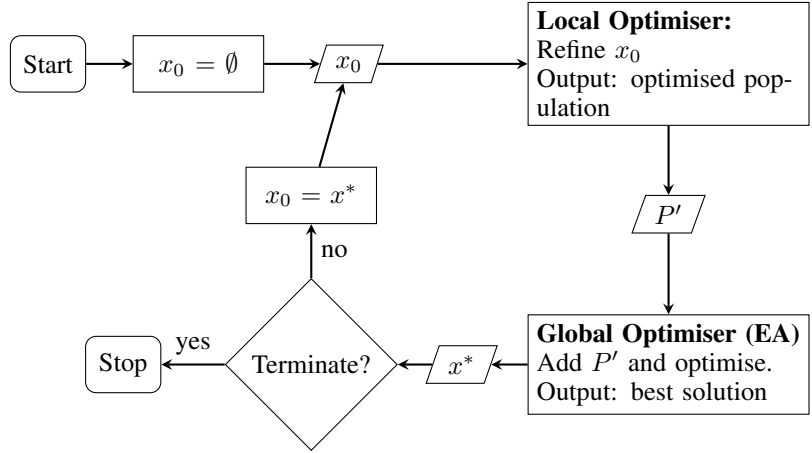 
\begin{algorithm}[!t]
    
    \begin{algorithmic}[1]
    \caption{(1+1) EA with local optimiser}\label{algo:1p1-mfo}
        \STATE $x^* \gets\{0\}^n$
        \WHILE{termination criterion is not satisfied}
            \STATE run local optimiser with $x^*$ as initial solution.
            \STATE $S_{local}\gets$ solution set from local optimiser after $eval_{local}$ evaluations.
            \STATE $x^* \gets \arg_x \max \sum_{i=1}^M P^{i}_{Cheb}$ (see Eq. \ref{eq:qmkp_kp_profit})
            \STATE run $(1+1)$ EA with $x^*$ as initial solution.
            \STATE $x^* \gets$ solution from $(1+1)$ EA after $eval_{global}$.
            \STATE 
        \ENDWHILE
    \end{algorithmic}
\end{algorithm}

\begin{algorithm}[!t]
    
    \begin{algorithmic}[1]
    \caption{$(\mu+\lambda)$ EA with local optimiser}\label{algo:muPlambda-mfo}
        \STATE $x^* \gets\{0\}^n$
        \WHILE{termination criterion is not satisfied}
            \STATE run local optimiser with $x^*$ as initial solution.
            \STATE $S_{local}\gets$ solution set from local optimiser after $eval_{local}$ evaluations.
            \STATE add $S_{local}$ to population in $(\mu+\lambda)$ EA and select best $\mu$ solutions.
            \STATE run $(\mu+\lambda)$ EA with for$eval_{global}$ evaluations.
            \STATE $S_{global}\gets$ solution set from $(\mu+\lambda)$ EA.
            \STATE $x^* \gets$ best solution in  $S_{global}$.
        \ENDWHILE
    \end{algorithmic}
\end{algorithm}

\subsection{Local Optimisation of Stochastic QMKP}

In this section, we propose a local optimiser for stochastic QMKP as a part of the hybrid evolutionary approach introduced above. This local optimiser is inspired by the composition of QMKP with multiple knapsacks. The problem consists of a set of knapsacks with individual weight capacity constraints and profit functions that collectively form the overall profit function. We aim to apply local optimisation for the problem via optimising the individual knapsack tasks.

\subsubsection{Localised Problem Model for Stochastic QMKP}\label{sec:local_model}

If each knapsack in QMKP can be considered as a separate optimisation task, one can use the multi-tasking optimisation techniques to solve the problem. However, this is not possible since the individual knapsacks in QMKP do not have disjoint search spaces. The search space of this problem is the set of all knapsack elements, each of which can be selected into any one of the knapsacks. To apply multi-tasking optimisation techniques, it is essential to consider a different problem model for local optimisation, in which each knapsack gets a disjoint subgroup of elements that can be assigned only to the respective knapsack. Therefore, we propose a localised model of the problem considering a reference solution, in which groups of the knapsack elements are formed as sub-search spaces for each knapsack. 

The goal of the local optimiser is to refine a solution provided by the EA. Therefore, each time a solution is received, the local optimiser generates new element groups. The input solution plays a key role during the process of grouping knapsack elements. During this process, for each item, the preferred knapsack, for which the item generates the highest value addition, is identified. The input solution is used as a reference for determining these knapsack preferences. The process of determining the knapsack preferences is as follows. 
For each item, its assignment is individually changed to each knapsack, while keeping the assignments of all other items in the reference solution unchanged. These different assignments are then evaluated to identify the knapsack in which the item contributes the highest profit addition. The items that prefer the same knapsack are grouped together, forming local search spaces for each knapsack. 

The preference of a particular knapsack element is determined based on its profit density in each knapsack. Let $x_0$ be the reference solution, then the profit density for each item $i$ for each knapsack $k\in M$ is calculated as follows, 
$$\text{profit-density}_i(x_0, k) = \frac{1}{w_i}\left({p_i + \sum_{i\neq j\in I_k} pij}\right)$$. 

This density value reflects the ratio between profit gain (or loss) and weight when the item $i$ in $x_0$ is assigned to (or removed from) knapsack $k$. Given the reference solution $x_0$, the above density indicates the change of the fitness value of $x_0$ if the item $i$ is assigned to or removed from knapsack $k$.

If the reference solution $x_0$ is an empty solution, the quadratic profit is $0$, and the profit density of item $i$ would be $p_i/w_i$ for all knapsacks. In this case, the items are sorted according to their profit density, and their preference is set for each knapsack $k \in M$ in a round-robin fashion. Otherwise, (when $x_0$ is not empty), the preference of item $i$ is set to the knapsack that gives the maximum profit density as follows, 
$$\text{arg}_k  \max_{k\in M} \text{ profit-density}_i\left(x_0, k\right).$$

The preference-based element groups create a localised problem model where the problem consists of multiple optimisation tasks. The set of optimisation tasks in this local problem can be described as follows. For each knapsack, we define individual optimisation tasks. This gives $m$ task fitness functions $g_1, \ldots, g_m$, such that $g_k$ is the fitness of knapsack $k\in M$. A fitness evaluation $g_k$ to evaluate fitness of the solution $x$ for each knapsack $k$ is derived from $g$ (see Equation \ref{eq:fitness-qmkp}) as follows,

\begin{equation}
    \label{eq:fitness-qmkp}
    g_k(x) = \left\{
    \begin{array}{lcl} 
       \sum_{i\in I_k} p_i + \sum_{i\neq j\in I_k} p_{ij} 
		 && \sum_{i\in I_k} w_i < C_k\\
         C_k - w(x)  && \text{otherwise}
    \end{array} \right.
\end{equation}

\subsubsection{Preference-Based Mutation for Localised Stochastic QMKP}\label{sec:local_mutation}

In addition to the problem model, it is important that the variation operators consider the disjoint search spaces when generating offspring solutions. The variation operators during the local optimisation require to comply with the disjoint subsearch spaces of the individual optimisation tasks (for each knapsack). We propose to use a simple mutation operator that considers these preference groups and generates offspring solutions for the local optimisation of QMKP. This mutation also considers the same solution representation used with EAs, a sequence of integers (see Section \ref{sec:representation_mutation}. 

Given an integer sequence $x=\{0,1, \ldots, m\}^n$, each integer in $x$ is mutated with a probability of $1/n$, similar to the random resetting mutation operator. If an item is decided to be mutated, its knapsack preference is considered as described in Algorithm \ref{alg:preference_mutation}. When mutating an item, if it is already assigned to the preferred knapsack, it is set to an unassigned state. Otherwise, it is set to the preferred knapsack.

\begin{algorithm}[!t]
    \begin{algorithmic}[1]
    \caption{Preference Based Mutation for Individual Items}\label{alg:preference_mutation}
    \REQUIRE $x$  (solution), $i$  (item to mutate) and $k^\prime_i$ (prefered knapsack for $i$)
    \IF{$i$ in $x$ = $k^\prime_i$} \STATE set $i$ is $x$ to $0$ (unassigned state).
    \ELSE \STATE  set $i$ of $x$ to $k^\prime_i$.
    \ENDIF
    \RETURN $P$
\end{algorithmic}
\end{algorithm}

\subsubsection{Multi-task Optimisation for  Localised Stochastic QMKP}\label{sec:mfo_method}

The local optimisation may consider the problem to consist of a set of quadratic (single) knapsack problems (QKPs). For each optimisation task, it is required to maximise the (item and pair) profit of assigned items and maintain a particular weight capacity. As these optimisation tasks have disjoint search spaces, the localised problem model allows us to apply multi-task optimisation techniques during local optimisation. 

We propose a local optimiser which follows the optimisation principles from Multi-Factorial Optimisation (MFO) to refine a given QMKP solution through local optimisation. MFO is a multi-task optimisation approach that has been used in evolutionary approaches, and it is particularly effective when the optimisation tasks share a similar problem structure \citep{MFEA_Gupta_7161358}. Since the optimisation tasks in the localised QMKP model are similar, MFO is the most promising approach to apply. We propose this MFO-based local optimisation as a part of the hybrid evolutionary approach to apply local refinements to solutions passed on from EAs. 

The optimisation with MFO depends on ranking solutions by their fitness for each task and a scalar fitness measure to unify the selection of solutions. MFO uses three metrics of a solution: factorial rank, skill factor and scalar fitness during the optimisation. We adopt these metrics for local optimisation of QMKP as follows. To model this as a multi-task optimisation problem, we consider $\{g_1, \ldots, g_m\}$ as the task fitness functions. Based on this, the following measures are considered for MFO on QMKP.

\begin{itemize}
	\item \textbf{factorial rank}($\rho_k(x)$):  the rank of the solution $x \in P$ according to fitness $g_k$
	\item \textbf{skill factor}($\tau(x)$): the knapsack for which $x$ gives the minimum factorial rank.
	$$\tau(x)= \arg_t \min\{\rho_t(x)\}$$
	\item \textbf{scalar fitness}($\phi(x)$): reciprocal of factorial rank of $x$ with respect to its skill factor $$\phi(x)= \frac{1}{\rho_{\tau(x)}{(x)}}$$
\end{itemize}

Algorithm \ref{alg:mfo} describes the MFO approach we propose as a local optimiser for QMKP.

\begin{algorithm}[!t]
    \begin{algorithmic}[1]
    \caption{MFO for Local Optimisation of QKMP}\label{alg:mfo}
    \REQUIRE $x_0$: the reference solution, $\{I^\prime_0,\ldots,I^\prime_m\}$: the subsets of knapsack items according to preferences of $x_0$
    \STATE $P \gets \{x_0\}$
    \FOR{$i=1$ to $\mu$}
        \STATE $P = P \cup \text{mutate}(x_0)$
    \ENDFOR
    \WHILE{!termination criteria}
        \STATE $P \gets P \cup \text{generate\_offsprings}(P,\{I^\prime_0,\ldots,I^\prime_m\})$
	\STATE Evaluate the skill factor and scalar fitness of all $x \in P$
	\STATE $P \gets $ sort $P$ by scalar fitness
	\STATE $P \gets$ select first $\mu$ solutions
    \ENDWHILE
    \RETURN $P$
\end{algorithmic}
\end{algorithm}

When generating solutions in Algorithm \ref{alg:mfo}, we mainly use the mutation operator introduced in Section \ref{alg:preference_mutation}. Additionally, we use a crossover operator, which plays a key role in MFO. In MFO, knowledge transfer between optimisation tasks is achieved via this crossover operator. The crossover operator is straightforward to use in evolutionary optimisation, as it combines two parent solutions to generate new solutions. As a knowledge transfer mechanism, the crossover combines two solutions that perform well for different optimisation tasks and generates solutions from them. Thus, the offspring inherit the knowledge from two different optimisation tasks. The probability of applying crossover in each iteration is determined by the algorithm parameter $Pr_{KT}$, the probability of knowledge transfer. We consider this probability as $Pr_{KT}=0.1$ in our investigations.

The knowledge transfer crossover requires two solutions, $x$ and $y$, which have different skill factors. First, the items in $x$ assigned a skilled task (knapsack) of $y$ are removed with a probability of $1/2$, and vice versa. Similarly, for each item $i$ in $x$ assigned to its skilled task (i.e., the preferred knapsack), we copy that value to item $i$ in $y$ with probability $1/2$ and vice versa. This crossover operation allows the transfer of knowledge between two differently skilled solutions, $x$ and $y$, and generates new offspring solutions.

\section{Experiments}\label{sec:experiments}

In our experimental analysis, we explore the effectiveness of the proposed methods on QMKP problem instances. The details of the selected problem instances and experimental settings are discussed in the following subsections. 

\subsection{Problem Instances}
The literature on QMKP usually adopts the benchmark data from quadratic knapsack problems (QKP) \citep{3heuristics_QMKP_10.1145/1143997.1144096,ExactMethod_QMKP_10.1287/ijoc.2018.0840}. For the experimental investigations in this paper, we consider the benchmark suite named QKPLIB from \citet{QKPLIB_data} and derive QMKP instances from it \citep{QKPLIB_Jovanovic2023,QKPLIB_data}. The instances in QKPLIB are characterised by the number of items, the relationship between item profits and weights, the relationship between the item weights and quadratic profit values and the density of the pairwise profits. 
In the experiments, we consider six instances with $ n=100, 500, 1000$ knapsack elements, where elements' profits and weights correlate weakly (instances labelled as weak-100, weak-500 and weak-1000) and strongly (instances labelled as weak-100, weak-500 and weak-1000). 
The quadratic profits in these instances also maintain a relation to the weights of the item pairs. For a given pair of items, the geometric mean of the two weights is considered the quadratic profit. For pair of items $i$ and $j$ with item profits $w_i$ and $w_j$ respectively the quadratic profit is determined as follows, $$p_{ij}=\sqrt{w_iw_j}.$$
This formulation ensures that the quadratic profits are related to the weights of the two items and effectively panalises the large differences between $w_i$ and $w_j$ \citep{QKPLIB_Jovanovic2023}. These instances have 25\% density of the pairwise profits, which means only 25\% of item pairs have been assigned a positive quadratic profit value, and the quadratic profit values of the remaining pairs have been set to $0$. Our experiments, the item profits and quadratic profit values in the benchmark instances are considered as the expected value of the respective profits.
  
Previous studies on deterministic QMKP, such as \citet{3heuristics_QMKP_10.1145/1143997.1144096} and \citet{ExactMethod_QMKP_10.1287/ijoc.2018.0840}, have adopted QMKP problem instances from benchmark instances that are originally designed for QKP (single knapsack variation). We follow the same approach to convert the QKP instances in the QKPLIB benchmark suite to QMKP instances \citep{3heuristics_QMKP_10.1145/1143997.1144096,ExactMethod_QMKP_10.1287/ijoc.2018.0840}. We assume that each knapsack in QMKP has the same capacity. Given that $m$ is the number of knapsacks, we consider each knapsack to have a capacity of 80\% of the total weight of knapsack elements to be equally divided between all knapsacks. 
Thus, the capacity of each knapsack $k=1,\ldots m$ is, 
$$C_1 = \ldots = C_m = \frac{ 0.8 \times \sum_i^n w_i}{m} $$

\subsection{Experimental settings}
 
We evaluate the methods across the following experimental setups. 
We run experiments with (1+1) EA and $(\mu+\lambda)$ EA using population size configurations of (20+10) EA. We conduct experiments using MFO for local improvements with both (1+1) EA and (20+10) EA. When combining MFO, we alternate between the EA and MFO optimisations every 500 fitness evaluations. We set the termination criteria to limit the optimisation to 5 million fitness evaluations or a wall time of 20 minutes, whichever occurs first. 
Other details of the algorithms, including the evolutionary operators used in EAs and MFO-based local optimisation, are clearly described in Section \ref{sec:methods}.
All experiments are conducted on a supercomputing cluster of Intel(R) Xeon(R) Platinum 8360Y CPU nodes. Each algorithm has been allocated a single CPU node with four cores and 1 GB of memory.

We run the experiments on problem instances with varying numbers of knapsacks, $m=3, 5$, and $10$. For the chance constraints settings, we consider the distribution of each profit to be $p_i \in \{\mu_i-\delta, \mu_i + \delta \}$ for $\delta = 25, 50$ and the chance constraint bounds to be $\alpha=0.90, 0.99$.  

For every experimental setting, we perform 30 independent runs of each algorithm. The results are summarised using mean and standard deviation (denoted as std in results tables), calculated based on the results from these 30 runs. Additionally, we conduct statistical significance testing using the Kruskal-Wallis method with Bonferroni correction, evaluating differences at a 95\% confidence level. These statistical comparisons are given under \'stat\' columns, where the methods are reffered to by the numbers assigned in the respective column headers. In each \'stat\' column, $X^{+}$ indicates a method performs significantly better than method $X$, while $X^{-}$ indicates it performs significantly worse than $X$. If there is no statistically significant difference between the current method and $X$, then $X$ does not appear. 


\subsection{Experimental Results}\label{sec:results}

\begin{table*}[!t]
    \fontsize{8pt}{9.6pt}\selectfont 
    \setlength{\tabcolsep}{1pt}
    \caption{Results for stochastic QMKP instances with 100 items}
    \label{tab:qmkp_results_100}
    \centering
    \begin{tabular}{|l|r|r|r|r r r|r r r|r r r|r r r|}
\hline

~ & ~ & ~& ~ & \multicolumn{3}{c|}{(1) (20+10) EA} & \multicolumn{3}{c|}{(2) (1+1) EA} & \multicolumn{3}{c|}{(3) (20+10) EA with MFO}  & \multicolumn{3}{c|}{(4) (1+1) EA with MFO}\\ 

~ & m& $\delta$ & $\alpha$ & mean & std & stat & mean & std & stat  & mean & std & stat & mean & std & stat  \\ \hline

\hline \multirow{6}{*}{\rotatebox[origin=c]{90}{weak-100}} & 3 & 25 & 0.90 & 	 59985.74 &	 911.49 &	 ~  &	\textbf{60129.74} &	 861.23 &	 ~  &	 59817.56 &	 845.53 &	 ~  &	 59993.08 &	 962.10 &	 ~  \\
~ & ~ & ~ & 0.99 & 	\textbf{54546.73} &	 783.85 &	 ~  &	 54349.42 &	 861.18 &	 ~  &	 54188.85 &	 767.43 &	 ~  &	 54232.54 &	 775.45 &	 ~  \\
~ & ~ & 50 & 0.90 & 	\textbf{57677.95} &	 893.29 &	 ~  &	 57504.46 &	 951.26 &	 ~  &	 57586.11 &	 760.64 &	 ~  &	 57448.02 &	 813.47 &	 ~  \\
~ & ~ & ~ & 0.99 & 	 46315.46 &	 775.99 &	 ~  &	\textbf{46380.49} &	 672.39 &	 ~  &	 46021.80 &	 857.79 &	 ~  &	 46013.74 &	 540.08 &	 ~  \\
~ & 5 & 25 & 0.90 & 	\textbf{45415.80} &	 647.39 &	 ~  &	 45051.73 &	 872.68 &	 ~  &	 44999.14 &	 718.15 &	 ~  &	 45255.03 &	 707.02 &	 ~  \\
~ & ~ & ~ & 0.99 & 	 39398.26 &	 648.08 &	 ~  &	\textbf{39673.09} &	 599.91 &	 ~  &	 39479.96 &	 707.60 &	 ~  &	 39449.87 &	 688.60 &	 ~  \\
~ & ~ & 50 & 0.90 & 	 42773.18 &	 654.82 &	 ~  &	 42701.39 &	 686.22 &	 ~  &	\textbf{42943.32} &	 593.02 &	 ~  &	 42584.90 &	 660.96 &	 ~  \\
~ & ~ & ~ & 0.99 & 	 29339.27 &	 2770.25 &	 $  3^{-} 4^{-}$ &	 28847.76 &	 2810.74 &	 $  3^{-} 4^{-}$ &	 31653.19 &	 596.99 &	 $1^{+} 2^{+}  $ &	\textbf{31721.13} &	 560.98 &	 $1^{+} 2^{+}  $ \\
~ & 10 & 25 & 0.90 & 	 30769.24 &	 481.19 &	 ~  &	 30927.05 &	 528.91 &	 ~  &	\textbf{30982.01} &	 576.26 &	 ~  &	 30860.49 &	 579.62 &	 ~  \\
~ & ~ & ~ & 0.99 & 	\textbf{25076.11} &	 464.50 &	 ~  &	 24979.44 &	 452.57 &	 ~  &	 24898.00 &	 546.77 &	 ~  &	 25006.17 &	 580.40 &	 ~  \\
~ & ~ & 50 & 0.90 & 	 28275.73 &	 409.16 &	 ~  &	 28246.53 &	 640.66 &	 ~  &	 28355.60 &	 483.08 &	 ~  &	\textbf{28361.74} &	 552.46 &	 ~  \\
~ & ~ & ~ & 0.99 & 	 11713.32 &	 1615.86 &	 $  3^{-} 4^{-}$ &	 11527.85 &	 1344.13 &	 $  3^{-} 4^{-}$ &	\textbf{17054.96} &	 657.58 &	 $1^{+} 2^{+}  $ &	 16716.54 &	 557.99 &	 $1^{+} 2^{+}  $ \\
\hline \multirow{6}{*}{\rotatebox[origin=c]{90}{strong-100}} & 3 & 25 & 0.90 & 	 49939.06 &	 735.92 &	 ~  &	 49622.25 &	 817.37 &	 ~  &	 49966.83 &	 737.27 &	 ~  &	\textbf{50055.25} &	 611.72 &	 ~  \\
~ & ~ & ~ & 0.99 & 	 44152.35 &	 558.96 &	 ~  &	 44180.50 &	 738.05 &	 ~  &	\textbf{44332.34} &	 574.83 &	 ~  &	 44145.05 &	 580.42 &	 ~  \\
~ & ~ & 50 & 0.90 & 	 47211.51 &	 635.28 &	 ~  &	\textbf{47513.97} &	 726.51 &	 ~  &	 47220.09 &	 593.48 &	 ~  &	 47299.04 &	 834.05 &	 ~  \\
~ & ~ & ~ & 0.99 & 	 23105.33 &	 6329.88 &	 $  3^{-} 4^{-}$ &	 22692.87 &	 7310.38 &	 $  3^{-} 4^{-}$ &	 36214.25 &	 570.02 &	 $1^{+} 2^{+}  $ &	\textbf{36377.82} &	 460.17 &	 $1^{+} 2^{+}  $ \\
~ & 5 & 25 & 0.90 & 	 37976.48 &	 560.33 &	 ~  &	\textbf{38093.21} &	 643.17 &	 ~  &	 38061.69 &	 725.69 &	 ~  &	 37868.96 &	 632.26 &	 ~  \\
~ & ~ & ~ & 0.99 & 	 32273.75 &	 576.32 &	 ~  &	 32378.14 &	 619.04 &	 ~  &	\textbf{32420.15} &	 518.39 &	 ~  &	 32316.20 &	 608.38 &	 ~  \\
~ & ~ & 50 & 0.90 & 	 35692.23 &	 603.73 &	 ~  &	\textbf{35782.19} &	 573.02 &	 ~  &	 35486.16 &	 589.49 &	 ~  &	 35619.12 &	 663.95 &	 ~  \\
~ & ~ & ~ & 0.99 & 	 6214.10 &	 3670.66 &	 $  3^{-} 4^{-}$ &	 6442.52 &	 3218.93 &	 $  3^{-} 4^{-}$ &	\textbf{23207.74} &	 1801.98 &	 $1^{+} 2^{+}  $ &	 22965.48 &	 2026.29 &	 $1^{+} 2^{+}  $ \\
~ & 10 & 25 & 0.90 & 	\textbf{26145.95} &	 396.14 &	 ~  &	 25898.21 &	 485.40 &	 ~  &	 26076.66 &	 480.12 &	 ~  &	 25931.16 &	 426.78 &	 ~  \\
~ & ~ & ~ & 0.99 & 	\textbf{20497.68} &	 431.21 &	 $   4^{+}$ &	 20214.35 &	 434.60 &	 ~  &	 20293.46 &	 391.78 &	 ~  &	 20041.39 &	 488.70 &	 $1^{-}   $ \\
~ & ~ & 50 & 0.90 & 	\textbf{23674.71} &	 464.38 &	 ~  &	 23336.47 &	 517.79 &	 ~  &	 23647.85 &	 520.20 &	 ~  &	 23459.68 &	 407.73 &	 ~  \\
~ & ~ & ~ & 0.99 & 	 107.71 &	 409.90 &	 $  3^{-} 4^{-}$ &	 281.25 &	 526.08 &	 $  3^{-} 4^{-}$ &	\textbf{10064.33} &	 775.17 &	 $1^{+} 2^{+}  $ &	 9686.72 &	 765.06 &	 $1^{+} 2^{+}  $

\\
\hline
\end{tabular}
\end{table*}

\begin{table*}[!t]
    \fontsize{8pt}{9.6pt}\selectfont 
    
    \setlength{\tabcolsep}{1pt}
    \caption{Results for stochastic QMKP instances with 500 items}
    \label{tab:qmkp_results_500}
    \centering
    \begin{tabular}{|l|r|r|r|r r r|r r r|r r r|r r r|}
\hline

~ & ~ & ~& ~ & \multicolumn{3}{c|}{(1) (20+10) EA} & \multicolumn{3}{c|}{(2) (1+1) EA} & \multicolumn{3}{c|}{(3) (20+10) EA with MFO}  & \multicolumn{3}{c|}{(4) (1+1) EA with MFO}\\ 

~ & m& $\delta$ & $\alpha$ & mean & std & stat & mean & std & stat  & mean & std & stat & mean & std & stat  \\ \hline

\hline \multirow{6}{*}{\rotatebox[origin=c]{90}{weak-500}} & 3 & 25 & 0.90 & 	\textbf{1722579.52} &	 8524.44 &	 $ 2^{+} 3^{+} 4^{+}$ &	 1713146.60 &	 8710.95 &	 $1^{-}   4^{+}$ &	 1711806.32 &	 9588.62 &	 $1^{-}   4^{+}$ &	 1695267.59 &	 8580.29 &	 $1^{-} 2^{-} 3^{-} $ \\
~ & ~ & ~ & 0.99 & 	\textbf{1692776.12} &	 8443.88 &	 $ 2^{+} 3^{+} 4^{+}$ &	 1679955.14 &	 9338.27 &	 $1^{-}   4^{+}$ &	 1682907.33 &	 10129.05 &	 $1^{-}   4^{+}$ &	 1669791.26 &	 8106.20 &	 $1^{-} 2^{-} 3^{-} $ \\
~ & ~ & 50 & 0.90 & 	\textbf{1709861.84} &	 9344.22 &	 $ 2^{+}  4^{+}$ &	 1697235.46 &	 10187.24 &	 $1^{-}   4^{+}$ &	 1702128.67 &	 9536.83 &	 $   4^{+}$ &	 1686571.62 &	 10748.54 &	 $1^{-} 2^{-} 3^{-} $ \\
~ & ~ & ~ & 0.99 & 	\textbf{1650192.63} &	 10373.69 &	 $ 2^{+}  4^{+}$ &	 1640119.20 &	 8144.00 &	 $1^{-}   4^{+}$ &	 1643749.07 &	 10032.75 &	 $   4^{+}$ &	 1628409.30 &	 9144.41 &	 $1^{-} 2^{-} 3^{-} $ \\
~ & 5 & 25 & 0.90 & 	\textbf{1175422.35} &	 8632.95 &	 $ 2^{+}  4^{+}$ &	 1163860.95 &	 7062.80 &	 $1^{-}   $ &	 1169718.35 &	 7177.50 &	 $   4^{+}$ &	 1160035.99 &	 7294.72 &	 $1^{-}  3^{-} $ \\
~ & ~ & ~ & 0.99 & 	\textbf{1146619.96} &	 7939.35 &	 $ 2^{+}  4^{+}$ &	 1133908.75 &	 9635.28 &	 $1^{-}  3^{-} $ &	 1142386.25 &	 7970.81 &	 $ 2^{+}  4^{+}$ &	 1133225.60 &	 9001.58 &	 $1^{-}  3^{-} $ \\
~ & ~ & 50 & 0.90 & 	\textbf{1162093.09} &	 7139.61 &	 $ 2^{+}  4^{+}$ &	 1148426.86 &	 7622.95 &	 $1^{-}  3^{-} $ &	 1160547.83 &	 7256.13 &	 $ 2^{+}  4^{+}$ &	 1150071.14 &	 8468.86 &	 $1^{-}  3^{-} $ \\
~ & ~ & ~ & 0.99 & 	\textbf{1104917.35} &	 8008.31 &	 $ 2^{+}  4^{+}$ &	 1092276.70 &	 8348.68 &	 $1^{-}  3^{-} $ &	 1102288.79 &	 6150.73 &	 $ 2^{+}  4^{+}$ &	 1090746.82 &	 9567.32 &	 $1^{-}  3^{-} $ \\
~ & 10 & 25 & 0.90 & 	\textbf{732749.10} &	 6813.02 &	 $ 2^{+} 3^{+} 4^{+}$ &	 727281.42 &	 6982.37 &	 $1^{-}   $ &	 727276.45 &	 4614.48 &	 $1^{-}   $ &	 725819.47 &	 5364.63 &	 $1^{-}   $ \\
~ & ~ & ~ & 0.99 & 	\textbf{701592.70} &	 4998.29 &	 $   4^{+}$ &	 700200.25 &	 5501.78 &	 ~  &	 698085.08 &	 5015.60 &	 ~  &	 698029.43 &	 4675.12 &	 $1^{-}   $ \\
~ & ~ & 50 & 0.90 & 	\textbf{718782.08} &	 4021.49 &	 $ 2^{+} 3^{+} 4^{+}$ &	 714591.56 &	 5565.07 &	 $1^{-}   $ &	 714934.12 &	 4832.68 &	 $1^{-}   $ &	 713968.35 &	 3825.46 &	 $1^{-}   $ \\
~ & ~ & ~ & 0.99 & 	\textbf{660596.15} &	 5553.61 &	 $ 2^{+}  4^{+}$ &	 656280.72 &	 5678.38 &	 $1^{-}   $ &	 657083.30 &	 5198.28 &	 ~  &	 656190.47 &	 4707.81 &	 $1^{-}   $ \\
\hline \multirow{6}{*}{\rotatebox[origin=c]{90}{strong-500}} & 3 & 25 & 0.90 & 	\textbf{4438036.21} &	 29111.10 &	 $ 2^{+}  4^{+}$ &	 4392490.73 &	 29906.76 &	 $1^{-}   4^{+}$ &	 4411863.41 &	 31241.46 &	 $   4^{+}$ &	 4356167.67 &	 38076.22 &	 $1^{-} 2^{-} 3^{-} $ \\
~ & ~ & ~ & 0.99 & 	\textbf{4415085.93} &	 21447.35 &	 $ 2^{+}  4^{+}$ &	 4363199.71 &	 29282.90 &	 $1^{-}  3^{-} $ &	 4394374.18 &	 27143.84 &	 $ 2^{+}  4^{+}$ &	 4341732.58 &	 30138.32 &	 $1^{-}  3^{-} $ \\
~ & ~ & 50 & 0.90 & 	\textbf{4423479.44} &	 28787.64 &	 $ 2^{+}  4^{+}$ &	 4386590.35 &	 32092.31 &	 $1^{-}   $ &	 4400927.03 &	 30943.84 &	 $   4^{+}$ &	 4364718.29 &	 30159.01 &	 $1^{-}  3^{-} $ \\
~ & ~ & ~ & 0.99 & 	\textbf{4377992.41} &	 25689.49 &	 $ 2^{+} 3^{+} 4^{+}$ &	 4322482.94 &	 25916.21 &	 $1^{-}   $ &	 4347462.50 &	 28325.89 &	 $1^{-}   4^{+}$ &	 4297091.08 &	 33402.04 &	 $1^{-}  3^{-} $ \\
~ & 5 & 25 & 0.90 & 	\textbf{3066607.52} &	 24534.51 &	 $ 2^{+} 3^{+} 4^{+}$ &	 3016000.37 &	 29264.22 &	 $1^{-}  3^{-} $ &	 3040555.98 &	 23427.34 &	 $1^{-} 2^{+}  $ &	 3024585.18 &	 23353.12 &	 $1^{-}   $ \\
~ & ~ & ~ & 0.99 & 	\textbf{3024047.40} &	 24844.78 &	 $ 2^{+}  4^{+}$ &	 2989280.79 &	 28113.40 &	 $1^{-}  3^{-} $ &	 3021520.56 &	 21615.27 &	 $ 2^{+}  4^{+}$ &	 2987078.40 &	 28735.61 &	 $1^{-}  3^{-} $ \\
~ & ~ & 50 & 0.90 & 	\textbf{3039316.49} &	 28458.20 &	 $ 2^{+}  4^{+}$ &	 3004062.22 &	 36140.59 &	 $1^{-}  3^{-} $ &	 3032554.10 &	 16342.04 &	 $ 2^{+}  $ &	 3017885.75 &	 21797.45 &	 $1^{-}   $ \\
~ & ~ & ~ & 0.99 & 	\textbf{2982165.95} &	 22428.31 &	 $ 2^{+}  4^{+}$ &	 2944682.90 &	 23997.87 &	 $1^{-}  3^{-} $ &	 2979326.14 &	 20514.57 &	 $ 2^{+}  4^{+}$ &	 2940726.07 &	 24911.08 &	 $1^{-}  3^{-} $ \\
~ & 10 & 25 & 0.90 & 	\textbf{1924014.93} &	 13976.23 &	 $ 2^{+} 3^{+} $ &	 1912886.79 &	 19747.29 &	 $1^{-}   $ &	 1910486.20 &	 13292.55 &	 $1^{-}   $ &	 1914504.14 &	 16090.47 &	 ~  \\
~ & ~ & ~ & 0.99 & 	\textbf{1894816.80} &	 15762.27 &	 $  3^{+} $ &	 1885536.62 &	 16930.78 &	 ~  &	 1882041.45 &	 19650.38 &	 $1^{-}   $ &	 1884064.84 &	 11784.49 &	 ~  \\
~ & ~ & 50 & 0.90 & 	\textbf{1906671.56} &	 17385.13 &	 ~  &	 1902241.12 &	 15812.70 &	 ~  &	 1903806.91 &	 16991.32 &	 ~  &	 1900784.48 &	 12144.29 &	 ~  \\
~ & ~ & ~ & 0.99 & 	\textbf{1848948.19} &	 15221.17 &	 $  3^{+} $ &	 1843634.06 &	 14825.97 &	 ~  &	 1837964.31 &	 13038.61 &	 $1^{-}   $ &	 1846261.61 &	 10175.11 &	 ~
\\
\hline
\end{tabular}
\end{table*}

\begin{table*}[!t]
    \fontsize{7.8pt}{9.36pt}\selectfont 
    \setlength{\tabcolsep}{1pt}
    \caption{Results for stochastic QMKP instances with 500 items}
    \label{tab:qmkp_results_1000}
    \centering
    \begin{tabular}{|l|r|r|r|r r r|r r r|r r r|r r r|}
\hline

~ & ~ & ~& ~ & \multicolumn{3}{c|}{(1) (20+10) EA} & \multicolumn{3}{c|}{(2) (1+1) EA} & \multicolumn{3}{c|}{(3) (20+10) EA with MFO}  & \multicolumn{3}{c|}{(4) (1+1) EA with MFO}\\ 

~ & m& $\delta$ & $\alpha$ & mean & std & stat & mean & std & stat  & mean & std & stat & mean & std & stat  \\ \hline

\hline \multirow{6}{*}{\rotatebox[origin=c]{90}{weak-1000}} & 3 & 25 & 0.90 & 	\textbf{4009230.45} &	 9062.32 &	 $ 2^{+} 3^{+} 4^{+}$ &	 3991855.21 &	 8377.20 &	 $1^{-}   $ &	 3989656.17 &	 8528.14 &	 $1^{-}   $ &	 3983707.54 &	 11009.89 &	 $1^{-}   $ \\
~ & ~ & ~ & 0.99 & 	\textbf{3954010.35} &	 8184.13 &	 $ 2^{+} 3^{+} 4^{+}$ &	 3934448.55 &	 9977.13 &	 $1^{-}   $ &	 3937102.15 &	 10036.32 &	 $1^{-}   4^{+}$ &	 3926845.34 &	 8779.06 &	 $1^{-}  3^{-} $ \\
~ & ~ & 50 & 0.90 & 	\textbf{3982637.81} &	 9832.72 &	 $ 2^{+} 3^{+} 4^{+}$ &	 3969666.25 &	 10048.91 &	 $1^{-}   $ &	 3972614.37 &	 12206.20 &	 $1^{-}   4^{+}$ &	 3962195.55 &	 11327.69 &	 $1^{-}  3^{-} $ \\
~ & ~ & ~ & 0.99 & 	 3828796.08 &	 226431.91 &	 $ 2^{+} 3^{+} 4^{+}$ &	 3809623.38 &	 230853.10 &	 $1^{-}   4^{+}$ &	\textbf{3856683.55} &	 10845.33 &	 $1^{-}   4^{+}$ &	 3838165.32 &	 10822.24 &	 $1^{-} 2^{-} 3^{-} $ \\
~ & 5 & 25 & 0.90 & 	 2641527.18 &	 7207.79 &	 $  3^{+} $ &	\textbf{2644457.80} &	 6438.29 &	 $  3^{+} 4^{+}$ &	 2628906.62 &	 7952.34 &	 $1^{-} 2^{-}  $ &	 2635397.97 &	 9567.63 &	 $ 2^{-}  $ \\
~ & ~ & ~ & 0.99 & 	 2587319.87 &	 8866.79 &	 $  3^{+} 4^{+}$ &	\textbf{2589425.56} &	 6796.69 &	 $  3^{+} 4^{+}$ &	 2570990.75 &	 7642.09 &	 $1^{-} 2^{-}  $ &	 2577847.84 &	 7905.94 &	 $1^{-} 2^{-}  $ \\
~ & ~ & 50 & 0.90 & 	\textbf{2618439.71} &	 9041.28 &	 $  3^{+} $ &	 2618293.34 &	 8002.76 &	 $  3^{+} 4^{+}$ &	 2603254.68 &	 8183.15 &	 $1^{-} 2^{-}  4^{-}$ &	 2611164.65 &	 8956.55 &	 $ 2^{-} 3^{+} $ \\
~ & ~ & ~ & 0.99 & 	 2086470.42 &	 140543.76 &	 $  3^{-} 4^{-}$ &	 2173196.06 &	 223492.61 &	 $  3^{-} 4^{-}$ &	 2494175.02 &	 13214.96 &	 $1^{+} 2^{+}  $ &	\textbf{2496106.20} &	 8244.24 &	 $1^{+} 2^{+}  $ \\
~ & 10 & 25 & 0.90 & 	 1543022.33 &	 5240.53 &	 $ 2^{-} 3^{+} $ &	\textbf{1560964.97} &	 6484.15 &	 $1^{+}  3^{+} 4^{+}$ &	 1530409.97 &	 5673.22 &	 $1^{-} 2^{-}  4^{-}$ &	 1548323.24 &	 4983.81 &	 $ 2^{-} 3^{+} $ \\
~ & ~ & ~ & 0.99 & 	 1486254.68 &	 5675.80 &	 $ 2^{-} 3^{+} 4^{-}$ &	\textbf{1503146.04} &	 6047.76 &	 $1^{+}  3^{+} 4^{+}$ &	 1474557.09 &	 5974.78 &	 $1^{-} 2^{-}  4^{-}$ &	 1492347.46 &	 4121.23 &	 $1^{+} 2^{-} 3^{+} $ \\
~ & ~ & 50 & 0.90 & 	 1519159.14 &	 7623.31 &	 $ 2^{-} 3^{+} $ &	\textbf{1534167.47} &	 5411.76 &	 $1^{+}  3^{+} 4^{+}$ &	 1507606.99 &	 6328.32 &	 $1^{-} 2^{-}  4^{-}$ &	 1523143.49 &	 5077.87 &	 $ 2^{-} 3^{+} $ \\
~ & ~ & ~ & 0.99 & 	 834473.13 &	 121005.79 &	 $  3^{-} 4^{-}$ &	 845298.39 &	 107556.18 &	 $  3^{-} 4^{-}$ &	\textbf{1382242.02} &	 44128.11 &	 $1^{+} 2^{+}  $ &	 1379057.75 &	 59279.12 &	 $1^{+} 2^{+}  $ \\
\hline \multirow{6}{*}{\rotatebox[origin=c]{90}{strong-1000}} & 3 & 25 & 0.90 & 	 3417188.12 &	 7349.64 &	 $  3^{+} $ &	\textbf{3421793.15} &	 6227.19 &	 $  3^{+} 4^{+}$ &	 3408355.28 &	 6517.90 &	 $1^{-} 2^{-}  4^{-}$ &	 3414385.03 &	 5645.90 &	 $ 2^{-} 3^{+} $ \\
~ & ~ & ~ & 0.99 & 	 3360606.43 &	 6852.63 &	 $ 2^{-} 3^{+} $ &	\textbf{3366363.09} &	 6668.93 &	 $1^{+}  3^{+} 4^{+}$ &	 3353318.55 &	 5740.04 &	 $1^{-} 2^{-}  4^{-}$ &	 3358789.58 &	 6227.68 &	 $ 2^{-} 3^{+} $ \\
~ & ~ & 50 & 0.90 & 	 3392260.56 &	 7242.39 &	 $ 2^{-} 3^{+} $ &	\textbf{3398952.82} &	 6213.18 &	 $1^{+}  3^{+} 4^{+}$ &	 3384572.23 &	 7014.59 &	 $1^{-} 2^{-}  4^{-}$ &	 3392161.90 &	 7080.72 &	 $ 2^{-} 3^{+} $ \\
~ & ~ & ~ & 0.99 & 	 1758942.74 &	 550304.86 &	 $  3^{-} 4^{-}$ &	 1680423.23 &	 679921.24 &	 $  3^{-} 4^{-}$ &	 3275416.80 &	 4848.36 &	 $1^{+} 2^{+}  $ &	\textbf{3281282.39} &	 8024.46 &	 $1^{+} 2^{+}  $ \\
~ & 5 & 25 & 0.90 & 	 2260604.68 &	 4780.69 &	 $ 2^{-} 3^{+} $ &	\textbf{2266975.47} &	 6200.90 &	 $1^{+}  3^{+} 4^{+}$ &	 2250207.27 &	 5727.95 &	 $1^{-} 2^{-}  4^{-}$ &	 2260778.29 &	 4942.59 &	 $ 2^{-} 3^{+} $ \\
~ & ~ & ~ & 0.99 & 	 2203025.44 &	 5711.48 &	 $ 2^{-} 3^{+} $ &	\textbf{2210594.10} &	 4838.39 &	 $1^{+}  3^{+} 4^{+}$ &	 2196402.34 &	 6780.35 &	 $1^{-} 2^{-}  $ &	 2201399.37 &	 6123.28 &	 $ 2^{-}  $ \\
~ & ~ & 50 & 0.90 & 	 2234495.44 &	 6197.60 &	 $ 2^{-} 3^{+} $ &	\textbf{2241518.37} &	 4188.01 &	 $1^{+}  3^{+} 4^{+}$ &	 2224543.88 &	 6497.95 &	 $1^{-} 2^{-}  4^{-}$ &	 2233989.45 &	 6674.12 &	 $ 2^{-} 3^{+} $ \\
~ & ~ & ~ & 0.99 & 	 635948.33 &	 318275.18 &	 $  3^{-} 4^{-}$ &	 650782.37 &	 319803.46 &	 $  3^{-} 4^{-}$ &	 1890912.65 &	 193862.77 &	 $1^{+} 2^{+}  $ &	\textbf{1977694.28} &	 193480.31 &	 $1^{+} 2^{+}  $ \\
~ & 10 & 25 & 0.90 & 	 1325734.94 &	 5484.96 &	 $ 2^{-}  4^{-}$ &	\textbf{1340270.83} &	 5150.35 &	 $1^{+}  3^{+} 4^{+}$ &	 1318681.04 &	 5786.04 &	 $ 2^{-}  4^{-}$ &	 1332273.64 &	 4640.13 &	 $1^{+} 2^{-} 3^{+} $ \\
~ & ~ & ~ & 0.99 & 	 1271884.96 &	 4353.19 &	 $ 2^{-} 3^{+} $ &	\textbf{1283825.22} &	 4492.30 &	 $1^{+}  3^{+} 4^{+}$ &	 1261721.95 &	 4211.29 &	 $1^{-} 2^{-}  4^{-}$ &	 1274031.65 &	 3823.43 &	 $ 2^{-} 3^{+} $ \\
~ & ~ & 50 & 0.90 & 	 1301698.65 &	 4396.39 &	 $ 2^{-} 3^{+} 4^{-}$ &	\textbf{1315805.19} &	 3974.20 &	 $1^{+}  3^{+} 4^{+}$ &	 1293841.11 &	 4702.28 &	 $1^{-} 2^{-}  4^{-}$ &	 1307660.82 &	 3200.09 &	 $1^{+} 2^{-} 3^{+} $ \\
~ & ~ & ~ & 0.99 & 	 72901.40 &	 93634.88 &	 $  3^{-} 4^{-}$ &	 47031.09 &	 62874.98 &	 $  3^{-} 4^{-}$ &	 887354.36 &	 76254.44 &	 $1^{+} 2^{+}  $ &	\textbf{909970.48} &	 76544.40 &	 $1^{+} 2^{+}  $

\\
\hline
\end{tabular}
\end{table*}

In this section, we discuss the results from the experimental investigations of the proposed methods and the comparative analysis with the general EA.

As described in Section \ref{sec:experiments},  we assume that both item profit and pairwise profits have an independent uniform distribution, each with a dispersion of $\delta=25, 50$. The result tables present a summary of results from 30 runs, and the profit of solutions is estimated using Equation \ref{eq:cc_qmkp}, which derives from Chebyshev's inequality. 

Table \ref{tab:qmkp_results_100} shows results for the experiments conducted on small problem instances with only 100 knapsack items. These instances are comparatively easy, and in most of the experimental runs, they were able to complete 5 million fitness evaluations within the allocated walltime limit. 
The results on weak-100 and strong-100 instances show that there is no significant difference between the (20+10) EA and (1+1) EA for smaller problem instances. For the case of three knapsacks ($m=3$) in these instances, EAs without MFO give results with better mean values. However, the standard deviation of these cases is lower when MFO is incorporated. As the number of knapsacks increases, the capacities of the individual knapsacks decrease, and the number of feasible solutions becomes more infrequent. In these cases (where $m=5,10$), MFO significantly improves the results, even for small instances such as weak-100 and strong-100. This improvement is quite significant in the strong-100 instance when considering $\delta=50$ and $\alpha=0.99$. In this case, the uncertainties are higher, and the expected reliability levels are also higher. Without MFO-based local optimisation, EAs are unable to identify feasible, high-quality solutions, subsequently producing solutions with low profit estimates. The statistical significance testing also reveals that the use of MFO-based local optimisation is effective when the capacity bounds are tight and uncertainties are higher.

Table \ref{tab:qmkp_results_500} shows results for instances with $500$ knapsack elements. For these instances, (20+10) EA shows better results over other algorithms in a considerable number of settings. That algorithm outperforms all other methods, in weak-500 instance with $m=3$ knapsacks, where the capacity bound is higher and uncertainties are lower ($\delta=25$). Additionally, (20+10) EA produces better results than other methods for weak-500 when the capacity bound is low (due to $m=10$), and the reliability level is low ($\alpha=0.90$). Comparatively, (1+1) EA produce lower results than (20+10) EA for most settings, for many settings of weak-500. Also, we observe that embedding MFO has been unable to improve results over respective algorithms. Instance strong-500 is considered harder to solve compared to weak-500 since the item profits in strong-500 are strongly correlated to their weights. Experiments on strong-500 with $m=3$ and $5$ show that (20+10) EA generates superior results in many settings compared to (1+1) EA with and without the MFO-based local optimiser. However, when there are more knapsacks ($m=10$), MFO helps improve the results of (1+1) EA and generates results comparable to (20+10) EA, as shown by statistical significance testing. 

Finally, Table \ref{tab:qmkp_results_1000} presents results for the problem instances with 1000 knapsack elements. As the number of elements increases, problem instances become harder to solve. The overall results for these instances show that (1+1) EA performs better than (20+10) instances in multiple settings. However, (20+10) EA produce better profits than (1+1) EA approaches, for weak-1000 with three knapsacks. As the number of elements increases, (1+1) EA produce better results for weak-1000. The results from EAs with MFO outperform other methods in several settings. MFO improves results for weak-1000 with 5 and 10 knapsacks for settings with $\delta=50$ and $\alpha=0.99$. 
Similarly, for the same uncertainty ($\delta=50$) and reliability ($\alpha=0.99$) levels, (1+1) EA with MFO produces significantly better results for strong-1000 than other methods.

In summary, the results show that proposed methods perform differently depending on the problem sizes, uncertainty level and capacity bounds. For instances of smaller and average size ($n=100$ and $500$), especially under lower uncertainty and looser capacity constraints, (20+10) EA produces better results. However, (1+1) EA generate superior results for larger instances, particularly when weight constraints are tighter and profits are distributed with higher variance. The integration of MFO with EAs demonstrates its benefits in scenarios where instances are under tight capacity bounds and high uncertainty, proving its effectiveness in guiding the search towards feasible, high-quality solutions. 

\section{Conclusions\label{sec:conclusion}}
In this paper, we investigated evolutionary approaches for the chance-constrained variation of QMKP, where profits are uncertain. QMKP is a complex problem where the fitness of a solution is evaluated against a non-linear function and multiple knapsack constraints. We explore two evolutionary methods for the problem, the population based (20+10) EA and single solution based (1+1) EA approaches. As a key contribution, we utilise a local problem model for QMKP as a multi-tasking problem and propose to use a local optimiser to address this multi-tasking aspect of QMKP. In this localised model, we consider an input solution and divide the knapsack elements into groups, limiting their assignment to a single preferred knapsack. This concept of localising QMKP according to a given solution was proposed for the first time. This effectively enabled the use of MFO as a local search optimiser to apply local refinements to an input solution. 
Our results show that (20+10) EA performs competitively on smaller instances with lower uncertainty, and (1+1) EA, especially when integrated with the proposed local optimiser, performs better on larger and more constrained instances. The use of MFO proves particularly beneficial in handling high uncertainty and tighter capacity bounds, as it enables the discovery of high-quality, feasible solutions where other methods often perform poorly. 
This work represents the first step in investigating the application of a multi-tasking model to a general optimisation problem via local optimisation. The methods developed here are extendable to other optimisation problems with a similar structure and properties. This study shows opportunities in designing adaptive local optimisation techniques for chance-constrained combinatorial problems.

\section*{Acknowledgements}
This work has been supported with supercomputing resources provided by the Phoenix HPC service at the University of Adelaide.

\bibliographystyle{ACM-Reference-Format}
\bibliography{main}


\end{document}